\documentclass[arxiv]{melba}

\usepackage{amsmath,amsfonts}
\usepackage{subcaption}
\usepackage{algorithm}
\usepackage{algpseudocode}
\usepackage{multirow}
\usepackage{bbding}
\usepackage{pifont}
\usepackage{makecell}
\usepackage{textcomp}  

\melbaid{2025:008}  
\doi{10.59275/j.melba.2025-4849}
\melbaauthors{Lyu, Gao and Staring}  
\email{\{d.lyu, r.gao\}@lumc.nl}

\volume{3}
\firstpageno{135}  
\melbayear{2025}  
\datesubmitted{2024-12-10}  
\datepublished{2025-05-12}  

\melbaspecialissue{Medical Imaging with Deep Learning (MIDL) 2020}
\melbaspecialissueeditors{Marleen de Bruijne, Tal Arbel, Ismail Ben Ayed, Hervé Lombaert}
\ShortHeadings{MCP-MedSAM}{Lyu, Gao and Staring}

\title{MCP-MedSAM: A Powerful Lightweight Medical Segment Anything Model Trained with a Single GPU in Just One Day}

\author{
  \name Donghang Lyu\aff{1*}\orcid{0009-0008-1446-9930},
  \name Ruochen Gao\aff{1*}\orcid{0000-0002-9411-3369},
  \name Marius Staring\aff{1}\orcid{0000-0003-2885-5812}
}

\affiliations{
	\num 1 \addr Division of Image Processing, Department of Radiology, Leiden University Medical Center, Leiden, the Netherlands \\	
    \num * \addr These authors contributed equally to this work
}

\abstract{Medical image segmentation involves partitioning medical images into meaningful regions, with a focus on identifying anatomical structures and lesions. It has broad applications in healthcare, and deep learning methods have enabled significant advancements in automating this process. Recently, the introduction of the Segmentation Anything Model (SAM), the first foundation model for segmentation task, has prompted researchers to adapt it for the medical domain to improve performance across various tasks. However, SAM's large model size and high GPU requirements hinder its scalability and development in the medical domain. To address these challenges, research has increasingly focused on lightweight adaptations of SAM to reduce its parameter count, enabling training with limited GPU resources while maintaining competitive segmentation performance. In this work, we propose MCP-MedSAM, a powerful and lightweight medical SAM model designed to be trainable on a single A100 GPU with 40GB of memory within one day while delivering superior segmentation performance. Recognizing the significant internal differences between modalities and the need for direct segmentation target information within bounding boxes, we introduce two kinds of prompts: the modality prompt and the content prompt. After passing through the prompt encoder, their embedding representations can further improve the segmentation performance by incorporating more relevant information without adding significant training overhead. Additionally, we adopt an effective modality-based data sampling strategy to address data imbalance between modalities, ensuring more balanced performance across all modalities. Our method was trained and evaluated using a large-scale challenge dataset, compared to top-ranking methods on the challenge leaderboard, MCP-MedSAM achieved superior performance while requiring only one day of training on a single GPU. The code is publicly available at \textcolor{blue}{\url{https://github.com/dong845/MCP-MedSAM}}.}


\begin{document}

\twocolumn[\maketitle]


\interfootnotelinepenalty=10000




\section{Introduction}

\begin{figure}[tb]
    \centering
    \includegraphics[scale=0.32]{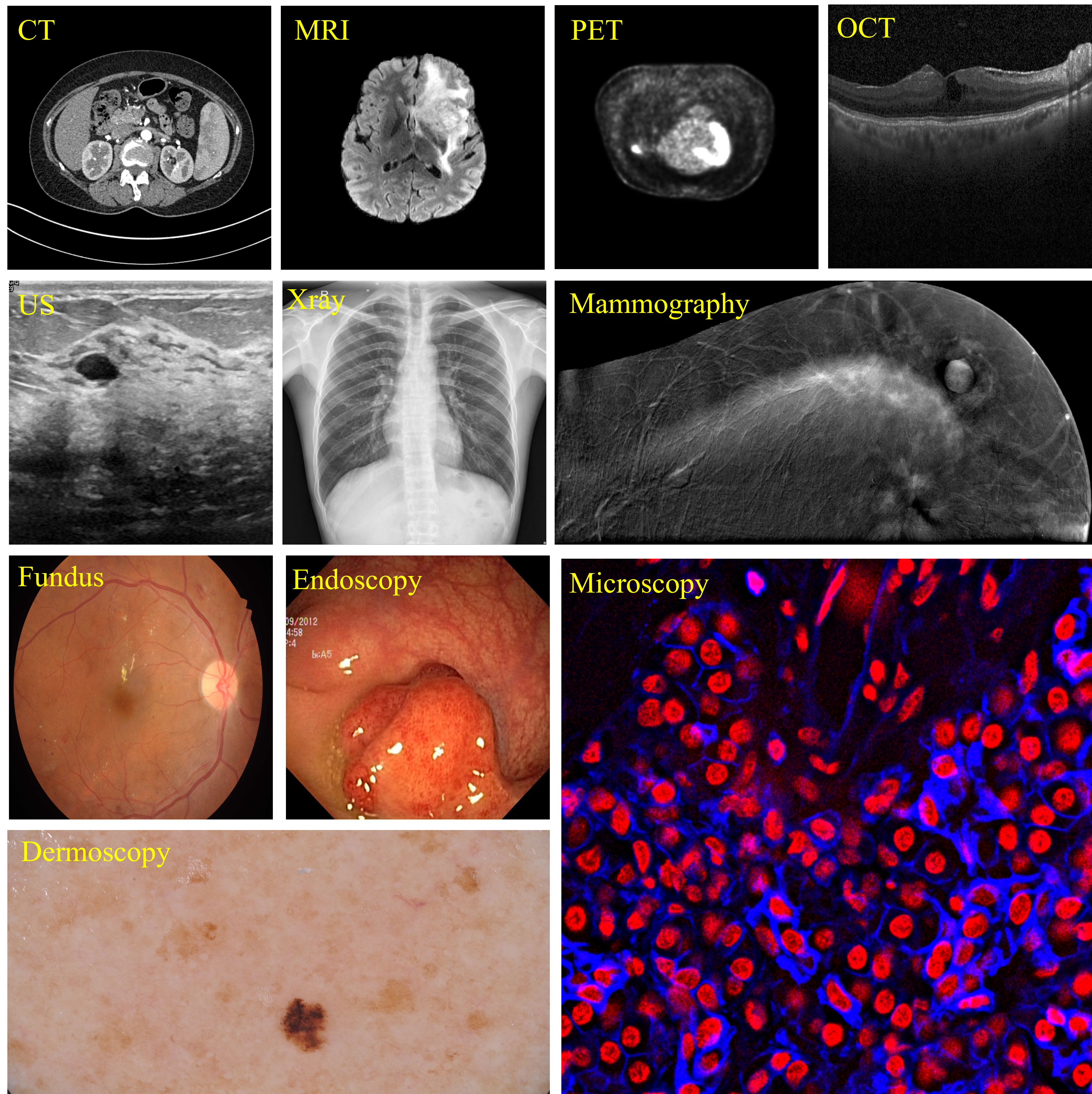}
    \caption{Examples of various medical imaging modalities.}
\label{fig:diverse_modalites}
\end{figure}

As a key task in the field of medical image analysis, medical image segmentation serves as the foundation for numerous clinical applications, such as disease diagnosis, treatment planning, and surgical interventions~\citep{litjens2017survey}. Accurate medical image segmentation can precisely identify anatomical structures and pathological regions, leading to more informed medical decisions. Over the last decade, deep learning-based segmentation models like nnU-Net ~\citep{isensee2021nnu} have achieved significant success in medical image segmentation.
These models are tailored to specific imaging modalities (e.g., MRI, CT, ultrasound, as shown in Figure~\ref{fig:diverse_modalites}) and particular diseases (e.g., prostate tumors, pulmonary nodules). This specialization means that different scenarios require different models. Additionally, their optimization for specific datasets limits their generalization across diverse data distributions~\citep{ma2024segment}.
Therefore, the development of a universal model for medical segmentation becomes increasingly promising in this context, offering the potential to unify approaches across various imaging modalities and clinical applications.

The introduction of the Segment Anything Model (SAM) ~\citep{kirillov2023segment}, the first foundation model for image segmentation, marked a significant breakthrough by offering a framework capable of generalizing across a wide variety of natural images. SAM’s robust generalization and zero-short capabilities have paved the way for the potential development of general medical segmentation models, enabling adaption across diverse medical image modalities while achieving strong performance on each. Building on SAM’s foundation, MedSAM~\citep{ma2024segment} was developed using a large-scale medical dataset covering 10 different imaging modalities and 30 types of diseases. This model has demonstrated superior overall performance compared to traditional, modality-specific specialist models, making a paradigm shift in medical image segmentation. By reducing the need to develop separate models for each imaging modality, it significantly streamlines the entire segmentation process.

Similar to many foundational models, SAM demands significant computational resources, including large-scale GPU clusters and long training time. These requirements limit the applicability and research on the SAM model, especially for smaller research groups and academic institutions that lack enough computational resources. Many previous studies on applying SAM to medical imaging concentrate on freezing the image encoder's weights for direct inference while adding supplementary components for adaptation~\citep{gao2023desam,zhang2023segment}, or introducing trainable adapters into the frozen image encoder~\citep{cheng2023sam,wu2025medical}. However, the simplicity of these introduced components limits their effectiveness, making it challenging for models to achieve optimal performance, especially when dealing with new imaging modalities or segmentation targets. Recently, some efforts have been made by LiteMedSAM\footnote{\url{https://github.com/bowang-lab/MedSAM/tree/LiteMedSAM}\label{junma}}, which distills the heavy image encoder into a lightweight version. This significantly reduces model parameters and curtails computational resource consumption, making MedSAM become more accessible for broader research and application. However, this efficiency comes at the expense of segmentation accuracy. Additionally, due to LiteMedSAM’s original training strategies, its performance is susceptible to imbalances in imaging modalities within the training data, further limiting its robustness across diverse medical imaging scenarios. Therefore, maintaining optimal overall performance of the MedSAM with a limited computational resource consumption still remains a critical goal. 

MedSAM's structure consists of three key components: an image encoder, a prompt encoder, and a mask decoder. While most prior research has primarily focused on optimizing the image encoder and mask decoder, the prompt encoder is usually left unchanged, typically relying on widely used prompt types such as boxes, points, and scribbles. Although these prompts offer valuable spatial hints to help the MedSAM segment targets more accurately, they still require manual effort for annotation, adding to the labor costs. Moreover, determining the optimal quantity and placement of points and scribbles within the given box remains challenging, impacting both model training and final segmentation performance~\citep{cheng2023sam1}. Therefore, optimizing the prompt encoder with some more effective and stable prompts could be beneficial, as it aims to enhance segmentation accuracy consistently without significantly increasing computational resources and labor costs. In this paper, we introduce \textbf{MCP-MedSAM}, which builds on the LiteMedSAM architecture to reduce computational resource costs and accelerate the training process. It incorporates two new types of prompts, a modality prompt and a content prompt, into the prompt encoder, generating sparse embedding and dense embedding representations and integrating them within the mask decoder to further improve segmentation accuracy. For the modality prompt, it is a learnable prompt and its embedding representation aims to take into account the differences among various imaging modalities, enriching the modality-specific information and helping the model adapt to diverse input characteristics. Meanwhile, the embedding representation of content prompt has two types: a sparse one and a dense one, both of which aim to leverage the direct information about the target within the given box but from different perspectives. Additionally, considering the strong zero-shot capability and extensive applications of the CLIP~\citep{radford2021learning} model, it is effectively integrated into the processing networks of both prompts to further enhance their representations. We also investigated various data sampling strategies and identified the most effective one to mitigate the negative impacts brought by the original imbalance. This resulted in enhanced overall segmentation performance, more balanced outcomes across modalities, and a faster training process. In summary, the key contributions of this work are shown as follows:
\begin{itemize}[left=0pt]
\setlength\itemsep{0em}

\item \textbf{Modality and Content Prompts}: We introduce two new types of prompts into the prompt encoder part of the MedSAM framework: the modality prompt and the content prompt. By generating effective embedding representations and integrating them with the mask decoder, the model's performance across all imaging modalities can be further enhanced without significantly increasing computational resource costs.

\item \textbf{Lightweight Architecture with Efficient Data Sampling Strategy}: We explore multiple data sampling strategies to identify the most efficient one, aiming to mitigate the effects of data imbalance and further accelerate the training process so that it converges within one day.

\item \textbf{Efficient Medical Segmentation Model for Broad Accessibility}: After evaluation on a large and diverse set of imaging data, our model demonstrates that it is feasible to achieve high-quality medical image segmentation without the need for extensive computational resources and long training time. This encourages more researchers to adopt and further explore general-purpose medical segmentation models.
\end{itemize}

\section{Related Work}

\subsection{Medical Image Segmentation}

Medical image segmentation has seen substantial advancements through the adoption of deep learning methods. The introduction of the U-Net model by \citet{ronneberger2015u} marks a significant milestone with its U-shaped architecture, effectively combining convolutional layers with symmetric contracting and expanding paths. Its widespread success inspires a host of variants aimed at boosting performance and tackling specific segmentation targets~\citep{isensee2021nnu, rahman2022deep, cai2022ma, fan2024csap,shu2024csca} by incorporating convolutional layers and attention mechanisms into the model design. Among them, nnU-Net~\citep{isensee2021nnu} stands out as a representative approach, leveraging optimized pre-processing and post-processing to achieve strong medical segmentation performance for each specific medical task. It has been widely adopted in various competitions and real-world applications. For instance, TotalSegmentator~\citep{wasserthal2023totalsegmentator, akinci2025totalsegmentator} offers comprehensive and practical solutions for multi-organ segmentation tasks in CT and MRI modalities.

Moreover, transformer modules have been widely integrated into U-shaped architectures for medical segmentation tasks, as transformers excel in extracting global contextual features via their self-attention mechanisms. Notable examples include UNETR~\citep{hatamizadeh2022unetr} and SwinUNETR~\citep{hatamizadeh2021swin}, which incorporate transformer modules into the encoder part of the U-Net architecture, yielding enhanced segmentation performance. Likewise, some works~\citep{chen2021transunet,chen2024transunet,tang2024htc} integrate CNN and transformer modules to leverage the strengths of both architectures and enhance segmentation performance. With the rise of Mamba~\citep{gu2023mamba}, recent works~\citep{ruan2024vm, wang2024mamba, liao2024lightm} have explored integrating the Mamba module into the U-Net architecture for further improvement.
Despite these advancements, the aforementioned models are generally tailored to specific segmentation tasks and exhibit limited generalizability across various medical imaging modalities.

\subsection{SAM}
SAM~\citep{kirillov2023segment} is an innovative model that aims to provide a versatile and generalizable solution for segmenting objects in images. There are two key concepts for the success of the SAM model: i) introducing multiple types of prompts, such as bounding boxes, points, and coarse masks, which allows the model to precisely identify and segment the target area; ii) training SAM with a huge amount of data, which enables SAM to adapt to a wide range of segmentation scenarios easily, reflecting its robust zero-shot capability. SAM has gained significant attention, leading to numerous recent works aimed at improving its performance and efficiency. HQ-SAM~\citep{ke2024segment} adopts a minimal adaptation approach by introducing a High-Quality output token and fusing global and local features from the image encoder to obtain high-quality features. In this way, HQ-SAM performs better on fine-grained segmentation tasks. SEEM~\citep{zou2024segment} introduces more kinds of prompts, including points, boxes, masks, scribbles and text prompts, and learns to deal with them by combining visual and text information in a joint visual-semantic space. This approach enhances segmentation performance and enables zero-shot adaptation to unseen user prompts. Then SAM2~\citep{ravi2024sam} extends the original SAM by enabling both image and video segmentation. By incorporating a memory mechanism, SAM2 can effectively process video data, leading to improved segmentation performance and broader applications. Although SAM models can achieve impressive segmentation performance, it is challenging to train without sufficient computational resources. SAM uses ViT-H~\citep{dosovitskiy2020image} as the image encoder and unifies the input image size to 1024 $\times$ 1024, both of which contribute to substantial GPU memory usage and make training become time-consuming. Therefore, there are also some works focusing on enhancing SAM's efficiency to make it more suitable for widespread real-world use. MobileSAM~\citep{zhang2023faster} distills the image encoder from ViT-H to a tiny ViT model, significantly reducing the parameter count. Furthermore, EfficientViT-SAM~\citep{zhang2024efficientvit} and RepViT-SAM ~\citep{wang2023repvit} replace ViT-H with some lightweight ViT variants, achieving better overall performance with significantly fewer parameters.

\subsection{Prior Knowledge in Medical Image Analysis}
For a model designed to handle multiple tasks, incorporating prior knowledge, such as features of anatomical structures or modality-specific characteristics, can enhance learning by helping the model recognize internal differences between tasks. DoDNet~\citep{zhang2021dodnet} employs one-hot embeddings to represent different organs, combining these embeddings with image features. This integrated representation is then fed into the output head, enabling DoDNet to segment tumors across various organs. Uniseg~\citep{ye2023uniseg} develops a learnable universal prompt that combines with sample-specific features to create prompts for multiple tasks. Task-specific prompts are then selected based on the task ID, allowing Uniseg to perform segmentation across multiple organs and modalities. Then MedPrompt~\citep{chen2024medprompt} introduces a self-adaptive prompt block designed to learn and incorporate cross-modal information, enhancing the model's ability to translate effectively across different modalities. Hermes model~\citep{gao2024training} introduces two kinds of learnable prompts, task-specific and modality-specific, which interact with the model in the bottleneck part to guide the model, enhancing segmentation performance across multiple organs and modalities. Prior knowledge has been effectively used in many models for multi-task processing but remains unexplored in the MedSAM framework for enhancing segmentation across various tasks.

\subsection{SAM for Medical Segmentation}
Inspired by the success of SAM, some works have begun exploring the utilization of SAM for medical segmentation tasks. MedSAM~\citep{ma2024segment} is proposed to adapt SAM for medical segmentation tasks by training it with a large dataset of medical images. Likewise, SAM-Med2D~\citep{cheng2023sam} fine-tunes the SAM model using a large-scale medical dataset and incorporates a variety of comprehensive prompts, including bounding boxes, points, and masks, rather than relying on just one type of prompt. Med-SA~\citep{wu2025medical} extend the SAM structure by introducing adaptors that highly enhance the capabilities for medical applications, enabling it to work with both 2D and 3D medical data. Furthermore, beyond the traditional prompts introduced by SAM, the ScribblePrompt~\citep{wong2024scribbleprompt} model explores the utilization of scribble prompts, resulting in improved overall performance. Similarly, SAT~\citep{zhao2023one} incorporates text prompts into the SAM model, offering contextual medical knowledge about modalities, anatomies, and body regions. With the release of SAM2, researchers have begun exploring its potential in medical segmentation, particularly its application for 3D medical segmentation. MedSAM-2~\citep{zhu2024medical} extends SAM2 for 2D and 3D medical segmentation by training on a large-scale medical dataset and incorporating a self-sorting memory bank to efficiently select informative embeddings, enhancing overall performance.

Furthermore, lightweight MedSAM models can be obtained by applying the techniques from lightweight SAM models and training them on medical data. MedficientSAM ~\citep{le2024medficientsam} distills the knowledge into an EfficientViT ~\citep{cai2023efficientvit} image encoder and fine-tunes it with a large-scale medical dataset, while Rep-MedSAM~\citep{wei2024rep} chooses to distill knowledge into a RepViT~\citep{wang2024repvit} image encoder. DAFT~\citep{pfefferle2024daft} also adopts the EfficientViT-SAM structure and introduces a data-aware fine-tuning policy inspired by the mixture of experts (MoE)~\citep{miller1996mixture} concept to further improve the performance of each modality. Swin-LiteMedSAM~\citep{gao2024swin} introduces a tiny Swin Transformer~\citep{liu2021swin} as the image encoder and builds skip connections between the image encoder and mask decoder. Furthermore, instead of solely using box prompt, box-based points and scribble prompts are automatically generated based on the provided bounding box. Then, LiteMedSAM-Rep~\citep{yang2024light} uses tiny Swin Transformer as the image encoder and trains the whole model from scratch, subsequently distilling the image encoder into a more lightweight RepViT while keeping the prompt encoder and mask decoder fixed. Although the above methods have made some progress in lightweighting the MedSAM, striking an optimal balance between training efficiency and performance remains a challenge.


\section{Materials and Methods}
\subsection{Datasets} 

\begin{figure}[tbp!]
    \centering
    \includegraphics[scale=0.5]{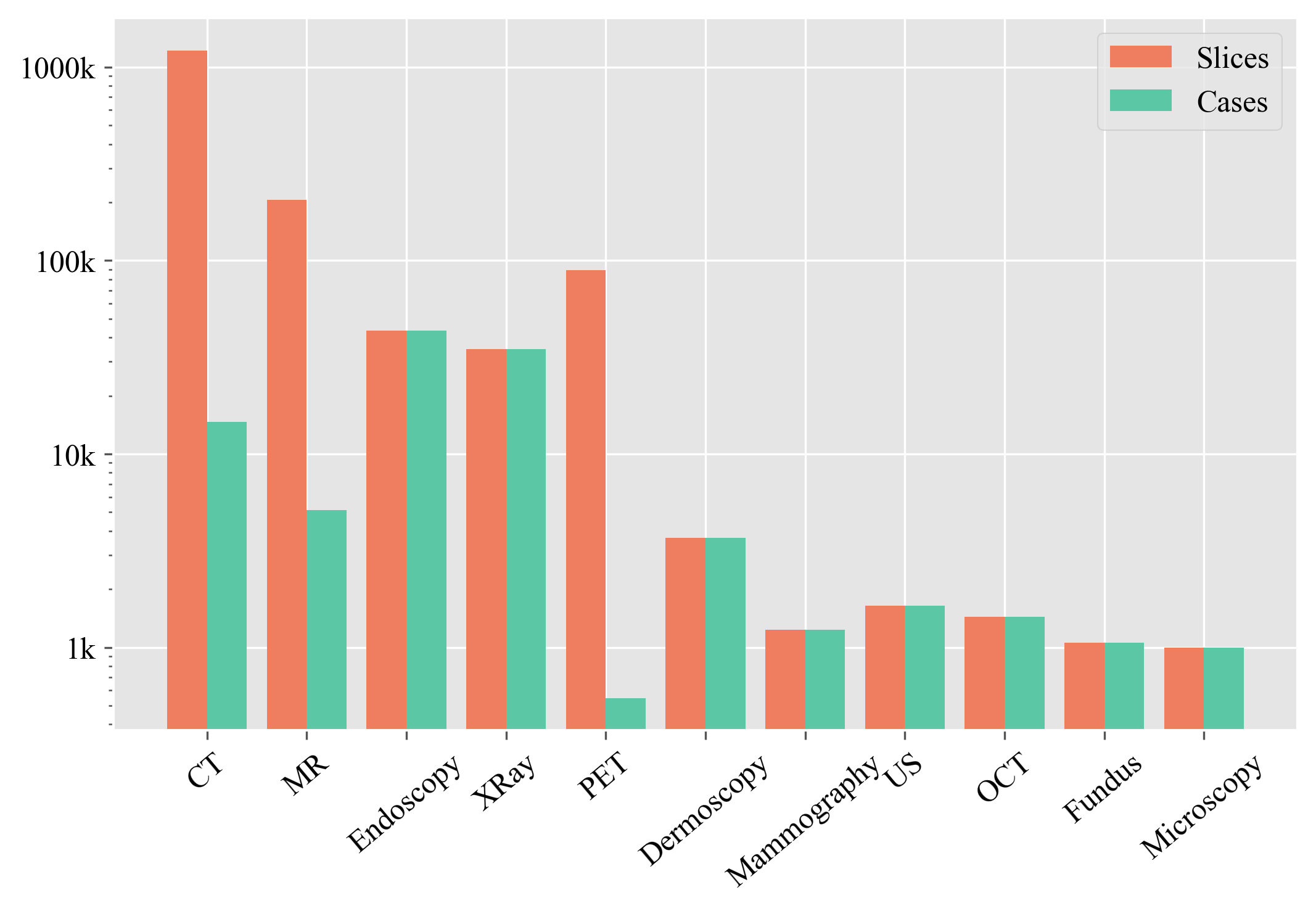}
    \caption{Data distribution across imaging modalities in the training set, with the y-axis displayed on a logarithmic scale for enhanced visualization.}
\label{fig:slice_case_distribution}
\end{figure}

For this study, we used the dataset~\citep{ma2024efficient} from the CVPR 2024 competition titled ``SEGMENT ANYTHING IN MEDICAL IMAGES ON LAPTOP''\footnote{\url{https://www.codabench.org/competitions/1847}\label{comp}} to train and test the model. The training dataset comprises over one million paired 2D images and their corresponding segmentation masks across 11 distinct imaging modalities, including Computed Tomography (CT), Magnetic Resonance Imaging (MRI), Positron Emission Tomography (PET), X-ray, Ultrasound, Mammography, Optical Coherence Tomography (OCT), Endoscopy, Fundus imaging, Dermoscopy, and Microscopy. The distribution of the training dataset across these modalities is illustrated in Figure~\ref{fig:slice_case_distribution}. As the testing set was not released after the challenge, we used the competition's validation set as our testing set. Then this testing set comprises 3,278 samples from 9 imaging modalities, excluding Mammography and OCT.

\subsection{MCP-MedSAM}
\begin{figure*}[tb]
    \centering
    \includegraphics[scale=0.32]{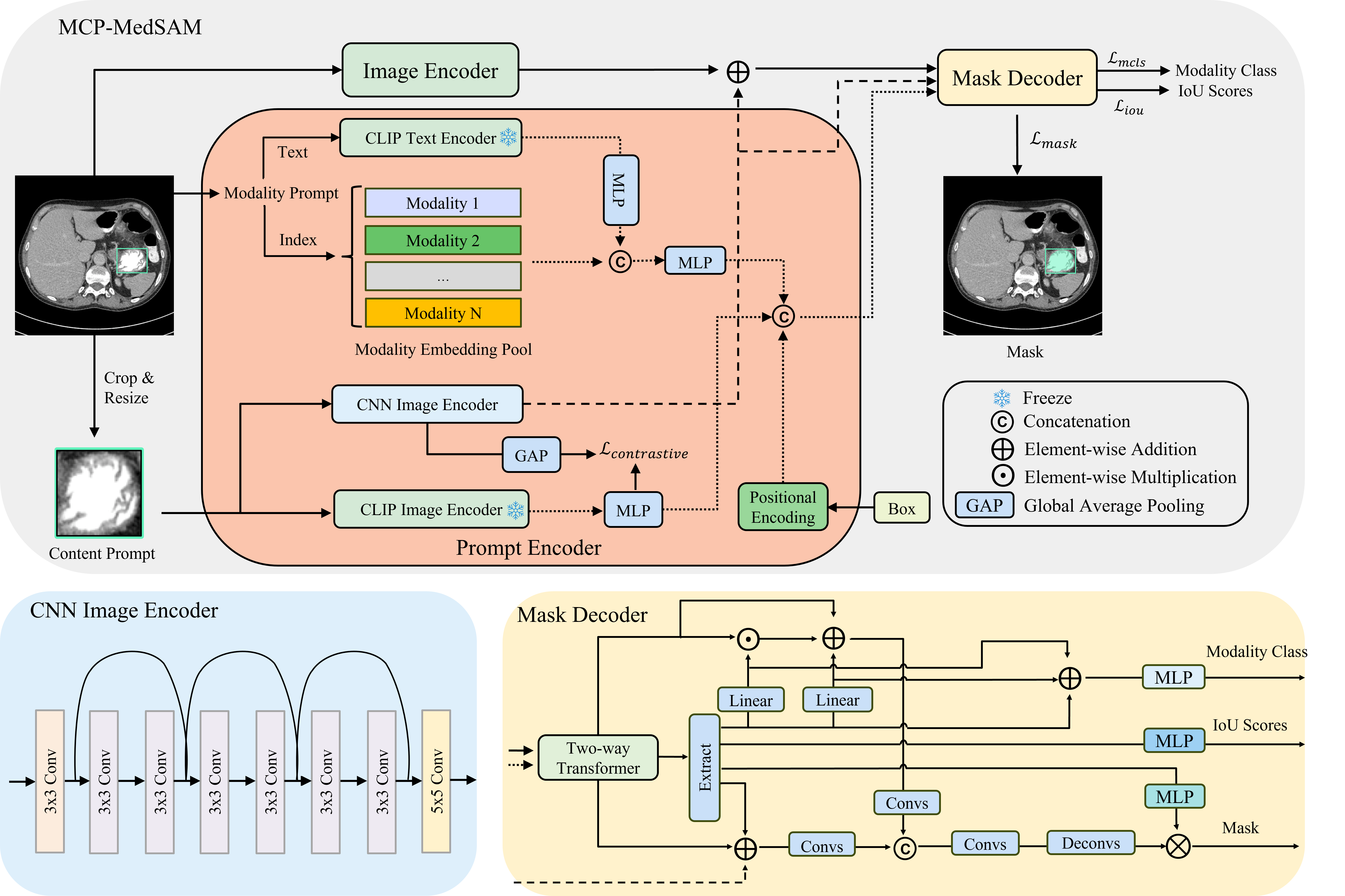}
    \caption{The top section provides an overview of MCP-MedSAM, highlighting our newly introduced content prompt and modality prompt in comparison to the SAM framework.  Notably, dotted lines indicate the flow of all sparse embedding representations, while dashed ones represent the direction of dense representation. The detailed architectures of the CNN Image Encoder and the modified Mask Decoder are illustrated in the bottom section.}
\label{fig:pipeline}
\end{figure*}

Overall, MCP-MedSAM follows the original design of SAM framework, which is composed of three parts: image encoder, prompt encoder and mask decoder, as shown in Figure~\ref{fig:pipeline}. We introduce two additional prompts into the prompt encoder and design a corresponding network to generate effective representations, enriching the output of prompt encoder. Additionally, the mask decoder is modified to better align with the representations of these prompts. Notably, similar to MedSAM, MCP-MedSAM is designed for 2D medical data and processes 3D medical data slice-by-slice.

While MedSAM-2 delivers strong segmentation performance, its training dataset is limited in both volume and modality diversity. MCP-MedSAM, in contrast, aims for faster convergence and broader modality support, requiring a robust pre-trained image encoder. To achieve this, we build on the lite version of MedSAM, which is trained on a larger, more diverse dataset spanning multiple modalities. A pre-trained tiny ViT\textsuperscript{\ref{junma}} image encoder from LiteMedSAM is used to accelerate convergence by providing some prior information. In addition to the traditional box prompt, we incorporate modality and content prompts into the prompt encoder to enhance relevance and robustness. The modality prompt consists of a modality text and an index for extracting the corresponding modality embedding, while the content prompt takes a cropped, resized image from the specified bounding box as input. Furthermore, the embedding representations of these two prompts are integrated into the mask decoder to better fuse modality and content information, resulting in improved segmentation performance. The mask decoder produces three outputs: the target mask, a predicted IoU score, and an additional prediction for the modality class.

\subsubsection{Modality Prompt}
Considering the unique characteristics of each modality, integrating modality information into LiteMedSAM is advantageous. Consequently, we introduce the modality prompt to enrich the sparse representation output by the prompt encoder. The modality prompt consists of two components: a modality text for generating a text embedding and a modality index $i$ for retrieving a modality-specific learnable embedding from the embedding pool $P \in \mathbb{R}^{N \times F}$. Here, $N$ is the number of modalities, $F$ denotes the embedding length, and modality index \( i \in \{1, \dots, N\} \) corresponds to a specific modality. In this context, we design each learnable embedding to encapsulate modality-specific information after training, while the text embedding provides a feature representation of the modality from a unique perspective. Given prior information of the modality class, we generate the corresponding text input, $\{Modality\}$ \textit{Image}, and pass it through the frozen CLIP text encoder to get text embedding. Although the CLIP model incorporates some medical-related prior information through its pre-trained weights, the training set contains a broader range of modalities and includes many unseen segmentation targets, adding complexity to the task. To address this, an MLP is added after the text encoder to tune and to adjust the channel dimensions simultaneously. On the other hand, a corresponding learnable modality-specific embedding is selected from the modality embedding pool based on the modality index. Then using an MLP to combine these two embeddings allows them to complement each other, creating a more complete and comprehensive modality representation that delivers modality-specific information effectively. The final modality embedding is appended to the sparse embedding. Notably, both MLPs in the modality prompt share the same structure, consisting of two linear layers with a GELU activation function. The input channel size is 512, while the output channel size is 256.



\subsubsection{Content Prompt}
The prompt encoder typically takes two types of input prompts: a sparse prompt and a dense prompt. The dense prompt serves as an initial coarse mask of the target, generating a dense embedding representation through a series of convolutional layers. When the dense prompt is unavailable, the dense embedding representation is instead modeled as a learnable matrix with weights initialized from a uniform distribution.  However, the learnable matrix is challenging to train effectively and lacks initial information about the target, while the coarse mask requires significant labor. Therefore, we introduce a content prompt to effectively capture target information within the specified box and generate a dense representation enriched with target features. Additionally, alongside the dense representation, we also extract a sparse embedding containing content information, aiming to enable more direct interaction with other sparse embeddings. Consequently, the final embedding representation of the content prompt has two components: a sparse one and a dense one, each capturing content information from different perspectives. By leveraging both components, the model can achieve a more comprehensive understanding of the content within the specified box, improving its ability to segment the target with greater accuracy. Given a bounding box, the content within the box is cropped and resized into a new shape first. The sparse content representation is derived by processing the reshaped input through a frozen CLIP image encoder, followed by an MLP with the same structure as the MLPs in the modality prompt part (two linear layers with GELU activation function, input channel size is 512 and output channel size is 256). This approach ensures that the sparse content embedding captures the comprehensive content information. For the dense content representation, the image is processed through a CNN image encoder based on the ResNet architecture~\citep{he2016deep}, consisting of multiple convolutional layers with skip connections. With an input channel size of 3, all subsequent convolutional layers maintain a consistent output channel size of 256. Given the small input size, this streamlined CNN network can efficiently extract local detailed features and preserve critical content information about the target. Since both the sparse and dense content embedding representations are generated from the same content, they should exhibit strong similarity. To achieve this, the dense content representation is passed through a global average pooling (GAP) layer to produce a sparse embedding. A contrastive loss is then applied between this sparse representation and the corresponding sparse content embedding to align their learning.

\subsubsection{Mask Decoder}
In the mask decoder, specific operations are employed to better fuse information from representations of both prompts, enhancing the final performance. After passing the sparse prompts and image embedding through the two-way transformer block, the updated modality embedding is extracted from its corresponding position of the sparse embeddings. In this context, two-way transformer block is a key component of the original SAM~\citep{kirillov2023segment} mask decoder. It employs self-attention on tokens and bidirectional cross-attention between tokens and image embeddings,  facilitating effective information exchange between image embeddings and tokens, which serve as sparse embeddings. These attention mechanisms also ensure the extracted modality embedding integrates both modality-specific and visual information. Then inspired by the thought of  FiLM~\citep{perez2018film}, we make the extracted modality embedding pass through two separate linear layers to produce weight and bias embeddings. These embeddings are subsequently combined with the dense matrix output from the two-way transformer through multiplication and addition operations, effectively incorporating modality-specific information. The combined matrix is further processed through two convolutional layers with a skip connection to join in the mask generation process. Furthermore, to better guide the network's learning of the modality prompt within the prompt encoder, we integrate a classification head into the mask decoder, following the same MLP structure as in the prompt encoder but with different input and output channels (two linear layers with GELU activation function, the input channel size is 256 and output channel size is the number of modalities, which is 11 in our experiments). This classification head takes the combination of the extracted embedding along with the corresponding weight and bias embeddings as input and predicts the corresponding modality class.

For the dense content representation, inspired by the global-local feature fusion in HQ-SAM~\citep{ke2024segment}, we combine it with the dense matrix output from the two-way transformer, which carries global visual information. Furthermore, the sparse content embedding is extracted from the sparse embedding output of the two-way transformer and integrated into this combination as a bias, further enriching the overall information. Finally, similar to the processing of modality information, the resulting matrix is processed through two convolutional layers with a skip connection to refine the features. We concatenate the outputs from both prompt sides and further process them through additional two convolutional layers with a skip connection to fuse the information. The resulting output is then upsampled twice to increase its size and is used to generate the target mask.  Notably, the MLPs used for predicting IoU scores and generating segmentation masks all share the same structure, consisting of three linear layers with ReLU activation functions. Both have an input channel size of 256, but their output channel sizes differ: the IoU prediction MLP outputs a single value, while the MLP for mask generation has an output channel size of 32 to align with the feature map.

\subsubsection{Loss Function}
In this work, except the traditional mask prediction loss $\mathcal{L}_{\text{mask}}$ and IoU (Intersection over Union) score prediction loss $\mathcal{L}_{\text{iou}}$, we also introduce another two loss components: $\mathcal{L}_{\text{mcls}}$ for the modality classification task and $\mathcal{L}_{\text{contrastive}}$ for approaching the similarity between two kinds of content prompts. Therefore, the overall loss function can be represented as follows:
\begin{equation}
    \mathcal{L} = \lambda_{1}\mathcal{L}_{\text{mask}}+\lambda_{2}\mathcal{L}_{\text{iou}}+\lambda_{3}\mathcal{L}_{\text{mcls}}+\lambda_{4}\mathcal{L}_{\text{contrastive}}.
\end{equation}
Here, all the $\lambda$ values are hyperparameters used to tune the importance of each loss component. We set $\lambda_{1}=\lambda_{2}=1$ and $\lambda_{3}=\lambda_{4}=0.01$, emphasizing the primary focus on segmentation while reflecting the auxiliary role of the classification and contrastive tasks in supporting overall performance.
\\
\\
\textbf{Mask Prediction Loss}  $L_{\text{mask}}$ is the sum of Binary Cross-Entropy (BCE) loss and Dice loss. Given the predicted mask $\hat{M}$ and ground truth mask $M$, BCE loss evaluates the pixel-wise difference, while Dice loss quantifies the overlap between the two masks. Then their working principles can be shown as follows:
\begin{equation}
    \mathcal{L}_{\text{BCE}} = - \sum_{i}^{N} \left( M_i \log(\hat{M}_i) + (1 - M_i) \log(1 - \hat{M}_i) \right),
\end{equation}
\begin{equation}
    \mathcal{L}_{\text{Dice}} = 1 - \frac{2 \sum_{i}^{N} M_i \hat{M}_i}{\sum_{i}^{N} M_{i}^{2} + \sum_{i}^{N} \hat{M}_{i}^{2}},
\end{equation}
where $N$ is the total number of pixels and $i$ is the pixel index. Then $\hat{M}_{i}$ and $M_{i}$ represent the \textit{i}-th pixel values of the predicted mask and the ground truth mask, respectively.
\\
\\
\textbf{IoU Loss}  $\mathcal{L}_{\text{iou}}$ measures the difference between the predicted IoU score from the model and the ground truth IoU score computed from the overlap between the predicted mask and the ground truth mask, aiming to further improve the accuracy of predicted segmentation mask. We employ a mean squared error (MSE) loss to capture this difference, encouraging precise IoU predictions:
\begin{equation}
    \mathcal{L}_{\text{iou}} = \frac{1}{N^{\prime}} \sum_{i=1}^{N^{\prime}} \left( s_{\text{iou}}^{i} - \hat{s}_{\text{iou}}^{i} \right)^2,
\end{equation}
where $N^{\prime}$ is the total number of masks, $s_{\text{iou}}^{i}$ is the ground truth IoU score for the $i$-th mask, while $\hat{s}_{\text{iou}}^{i}$ indicates the corresponding predicted IoU score.
\\
\\
\textbf{Modality Classification Loss} To ensure that modality prompt effectively represents modality-specific information, we introduce an auxiliary modality classification task by employing cross entropy loss function, which is defined as follows:
\begin{equation}
    \mathcal{L}_{\text{mcls}} = - \sum_{i=1}^{C} y_i \log(\hat{y}_i),
\end{equation}
where $C$ denotes the total number of modalities, $y_{i}$ is the label for each modality class $i$ and $\hat{y}_i$ is the predicted probability for each modality class $i$.
\\
\\
\textbf{Contrastive Loss} To ensure that pairs of sparse and dense content prompts exhibit the highest similarity, we introduce an auxiliary task that utilizes the contrastive loss from the CLIP~\citep{radford2021learning} method. The overall process can be represented as follows:
\begin{equation}
    \text{sim}_1 = F_{dc} \cdot F_{sc}^{T},
\end{equation}
\begin{equation}
    \text{sim}_2 = F_{sc} \cdot F_{dc}^{T},
\end{equation}
\begin{equation}
    \mathcal{L}_{\text{contrastive}} = \left( \mathcal{L}_{\text{ce}}(\text{sim}_1, \mathbf{y}) + \mathcal{L}_{\text{ce}}(\text{sim}_2, \mathbf{y}) \right)/2.
\end{equation}
Here, $F_{dc}$ and $F_{sc}$ are the normalized embeddings of the dense and sparse content prompts, respectively, both having the shape $\mathbb{R}^{B \times C}$, where $B$ is batch size and $C$ denotes the embedding length. $\text{sim}_{1}$ and $\text{sim}_{2}$ are two similarity matrices for these two embeddings, each with a shape of $\mathbb{R}^{B \times B}$, where $\text{sim}_{2}$ is the transpose of $\text{sim}_{1}$. $\mathcal{L}_{\text{ce}}$ denotes the cross entropy loss function and $\mathbf{y}$ are the collection of labels.

\subsection{Training Strategy}
In this section, we detail two core components of our training strategy: pre-trained modules (tiny ViT and CLIP) and data sampling strategy. Both are important for accelerating the model's convergence and obtaining a superior overall performance.

\subsubsection{Pre-trained modules}
As previously mentioned, many methods freeze pre-trained weights from SAM's image encoder, fine-tuning newly introduced learnable components to improve adaptability of new tasks. Using these pre-trained weights ensures a strong baseline ability for the model. However, instead of following this approach, we opt for a lightweight image encoder, tiny ViT, and use pre-trained weights specifically tailored for the medical image. This allows the image encoder of MCP-MedSAM to fully participate in the training process without being frozen, achieving stronger performance. Additionally, we utilize the pre-trained PubMedCLIP\footnote{\url{https://huggingface.co/kaushalya/medclip}}~\citep{eslami2021does} as the frozen CLIP component of MCP-MedSAM, aiming to leverage its strong zero-shot capability and make it provide some medical domain related prior information as well. This variant of CLIP is fine-tuned for the medical domain using the Radiology Objects in Context (ROCO) dataset~\citep{pelka2018radiology}, including multiple modalities from various human regions.

\subsubsection{Data sampling strategy}
As shown in Figure \ref{fig:slice_case_distribution}, the distribution of different image modalities in our training dataset is highly imbalanced, primarily due to two factors.

The first factor is varying sizes of public datasets: some modalities, such as CT, MR, and X-ray, have much more publicly available data for AI tasks, resulting in a significantly larger number of training samples. The second factor relates to data dimension. Modalities like CT and MR are in 3D format, enabling the extraction of significantly more slices compared to 2D modalities such as Dermoscopy, Fundus, and Microscopy. 

In SAM~\citep{kirillov2023segment} and MedSAM~\citep{ma2024segment}, data sampling involves iterating over all image slices. In our training set, this approach leads to a significant imbalance: CT slices account for approximately 76\% of the dataset, MR slices nearly 13\%, while Microscopy slices represent less than 0.1\%, making it the least represented modality. Such severe imbalance would negatively impact the model's performance, while training on a large number of slices also results in extended training time. 

To address the limitations of existing data sampling methods, we implement a modality-based data sampling strategy, a variant of stratified sampling. This approach prioritizes achieving balance across all modalities, as determined through comparisons with other commonly used data sampling strategies. The details of this strategy are outlined in Algorithm \ref{alg:sampling}. The key point of this approach is randomly selecting a slice from each data case and ensuring that all modalities are evenly sampled within each training batch. This method alleviates the negative impacts of severe data imbalance, allowing all modalities to be trained with an approximately equal number of slices.

\begin{algorithm}[tb]
\caption{Modality-based Data Sampling Strategy}
\label{alg:sampling}
\begin{algorithmic}[1]
\State \textbf{Input:} Total number of modalities $N$, number of cases in each modality $\{C_j\}_{j=1}^{N}$, batch size $B$
\State Initialize batch $\mathcal{B} = \emptyset$
\For {each epoch}
    \For {each batch of size $B$}
        \For {each sample in the batch}
            \State Select modality $m \sim \mathcal{U}\{0, N-1\}$
            \State Select case from the chosen modality: $c \sim \mathcal{U}\{0, C_m-1\}$
            \If {the selected case is 3D}
                \State Let the shape of the 3D case be $[Z, H, W]$
                \State Choose slice index $z \sim \mathcal{U}\{0, Z-1\}$
                \State Extract the slice $\mathcal{S} = \mathcal{D}_{m,c}[z, :, :] \in \mathbb{R}^{H \times W}$
            \Else
                \State Extract the 2D case $\mathcal{S} = \mathcal{D}_{m,c} \in \mathbb{R}^{H \times W}$
            \EndIf
            \State Select mask $k \sim \mathcal{U}\{0, K-1\}$, where $K$ is the number of masks in $\mathcal{S}$
            \State $\mathcal{B} \leftarrow \mathcal{B} \cup \{(\mathcal{S}, k)\}$
        \EndFor
    \EndFor
\EndFor
\State \textbf{Output:} Batch $\mathcal{B}$
\end{algorithmic}
\end{algorithm}

\section{Experiments and Results}

\subsection{Experimental Setup}
All experiments were performed using Python 3.10 and PyTorch 2.0.0 on a single NVIDIA A100 GPU with 40GB of memory. The AdamW optimizer was used with an initial learning rate of $2 \times 10^{-4}$ and a weight decay of $1 \times 10^{-3}$. During training, the learning rate was reduced by a factor of 0.9 every 5 epochs. The batch size was set to 16, and training was conducted for a total of 25 epochs. In the final epoch, we set a small learning rate to $5 \times 10^{-5}$ for an additional fine-tuning of the model. Data augmentation included vertical and horizontal flips, each applied with a 50\% probability. Additionally, the Wilcoxon signed-rank test is used to assess the statistical significance of the proposed method in the experiments.

\subsection{Evaluation Metrics} 
\textbf{Accuracy Metrics} Following \cite{ma2024segment}, we adopted the Dice Similarity Coefficient (DSC) and Normalized Surface Dice (NSD) as evaluation metrics. DSC measures the overlap between two sets, quantifying the similarity between the predicted segmentation and the ground truth. In contrast, NSD evaluates how closely the surfaces of the predicted segmentation align with the ground truth, focusing on surface accuracy. Higher DSC and NSD values correspond to greater segmentation accuracy. To note,  DSC and NSD are first averaged within each modality, and then these modality-wise means are averaged to obtain the overall value. Consequently, the standard deviation is also calculated at the modality level.
\\
\\
\textbf{Efficiency Metrics}
In order to evaluate the computational efficiency of the model, we established two metrics: GPU training time and CPU inference time. GPU training time is primarily used to assess GPU resource consumption, while CPU inference time focuses on evaluating model performance on edge devices without GPU support, such as laptops or standard CPU workstations. CPU inference time is also part of the competition evaluation criteria\textsuperscript{\ref{comp}} for the testing set. The CPU inference time was measured on our local platform with an Intel Xeon(R) W3-2435 @ 3.1 GHz processor and 8GB of memory, where the average time per case was tested in a Docker environment provided by the authors.

\subsection{Results}
\subsubsection{Comparison with Benchmark Models}
\begin{figure*}[tb]
    \centering
    \includegraphics[scale=0.25]{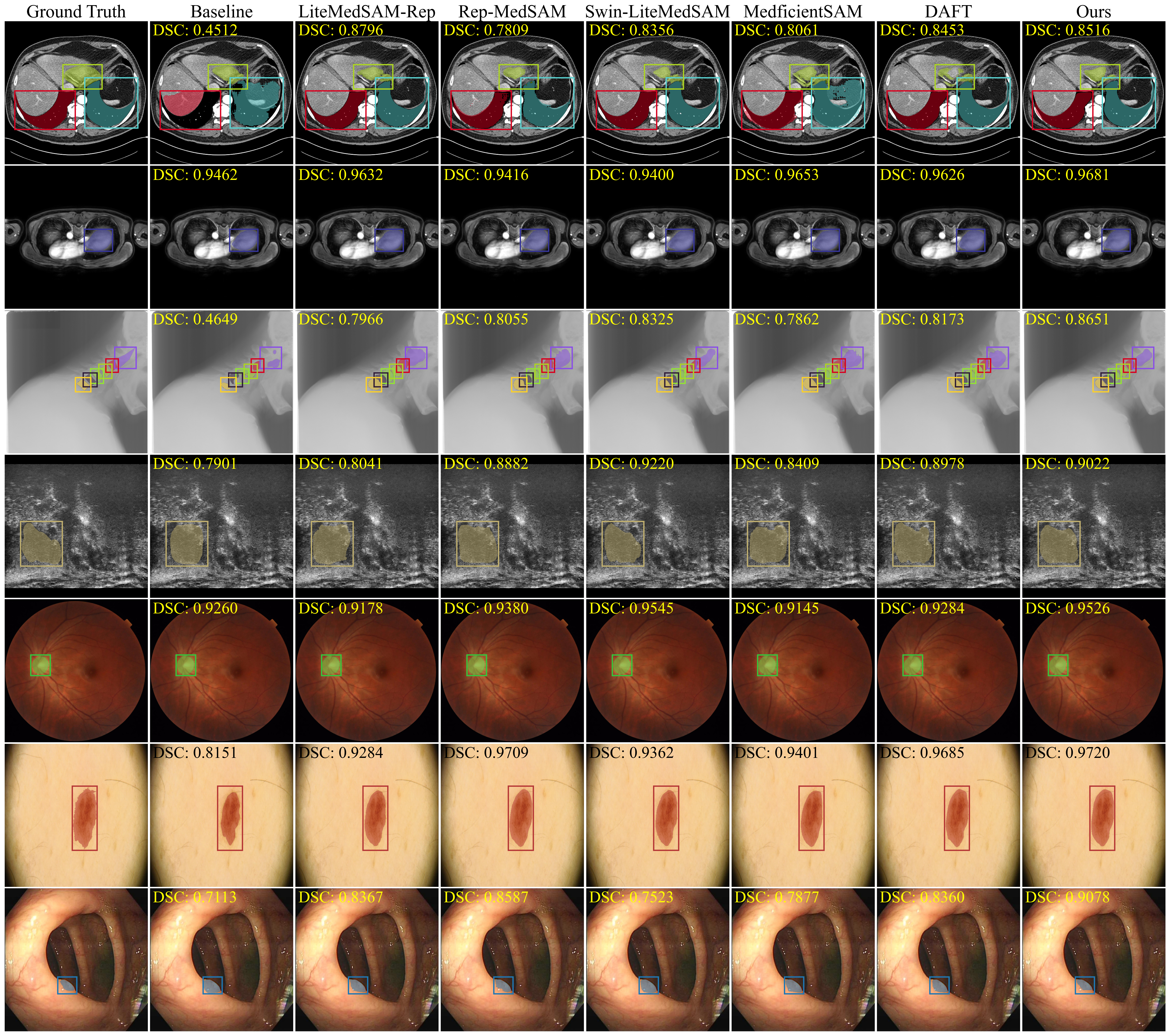}
    \caption{Visualization of multiple modalities yielded by our proposed method and the other benchmark models. The DSC value of each sample is displayed in the upper left corner.}
\label{fig:leaderboard}
\end{figure*}

First, we used the lightweight MedSAM model released by the organizers of the "SEGMENT ANYTHING IN MEDICAL IMAGES ON LAPTOP" competition at CVPR 2024 as our baseline. This model was developed by distilling the image encoder of MedSAM ~\citep{ma2024segment} and fine-tuning both the encoder and decoder together using the challenge dataset. Then we compared our model with the top-ranking models from the competition leaderboard: (1) MedificientSAM~\citep{le2024medficientsam}: MedificientSAM utilizes EfficientViT as its image encoder and increases the input size from 256 $\times$ 256 to 512 $\times$ 512; (2) DAFT~\citep{pfefferle2024daft}: DAFT also employs EfficientViT, fine-tuning it on different data subsets based on modality to generate multiple modality-specific models; (3) Rep-MedSAM~\citep{wei2024rep}: Rep-MedSAM employs RepViT as its image encoder and, after distillation, trains the entire model using the dataset; (4) Swin-LiteMedSAM~\citep{gao2024swin}: Swin-LiteMedSAM utilizes the tiny Swin Transformer as its image encoder and introduces point and scribble prompts, which are automatically generated based on bounding boxes. (5) LiteMedSAM-Rep~\citep{yang2024light}: LiteMedSAM-Rep's prompt encoder and mask decoder are initially trained from scratch alongside a tiny Swin Transformer as the image encoder. Then they distill the knowledge from the tiny Swin Transformer to RepViT. Overall, these models mainly focus on modifying the image encoder by distilling knowledge from a large ViT into a lightweight transformer encoder. Table~\ref{tab:Leaderboard} summarizes the segmentation performance of all models, reported in terms of DSC (\%) and NSD (\%). MCP-MedSAM achieved the best DSC and NSD performance compared to other benchmark models on the testing set. Notably, the segmentation predictions of these benchmark models were reproduced using Docker images provided by the authors, available on Docker Hub\footnote{\url{https://hub.docker.com/}}. Then Figure~\ref{fig:leaderboard} presents visualizations of the models' predictions, providing additional details. Overall, our proposed MCP-MedSAM produced segmentation masks that closely resembled the ground truth, whereas some other models exhibited either a higher degree of over-segmentation or of under-segmentation.

Table~\ref{tab:Leaderboard_time} demonstrates GPU training time (in hours) and CPU inference time (in seconds) across different methods. MCP-MedSAM achieved the shortest GPU training time (23.8 hours), significantly outperforming all other models. GPU training times were sourced from the respective papers, while the baseline model's training and inference times were not disclosed by the challenge organizers. To note, DAFT has three training stages, but it does not specify the time required for its third stage. Therefore, we estimated the total training time to exceed the combined duration of the first two stages. Additionally, LiteMedSAM-Rep and Rep-MedSAM utilized different types of GPUs (NVIDIA RTX 4090 and NVDIA V100). However, based on Lambda GPU benchmark analysis\footnote{\url{https://lambda.ai/gpu-benchmarks}}, we find that the A100 40GB is slightly faster than the RTX 4090 (by approximately 1.2×) and significantly faster than the V100 (by about 3.6×). Hence, we estimate that the equivalent training time on an A100 would be approximately 1333 hours for LiteMedSAM-Rep and 52 hours for Rep-MedSAM. As MCP-MedSAM requires only one day (23.8 hours) for training, and LiteMedSAM-Rep and Rep-MedSAM would take an estimated 56 and 2.2 days respectively under the same type of GPU, we can still conclude that MCP-MedSAM requires the least training time. For CPU inference time, most methods achieved approximately 1 second. MCP-MedSAM was the slowest among the compared methods, whereas DAFT was the fastest.


\begin{table*}[tb]
\caption{Accuracy comparison with state-of-the-art methods on the challenge leaderboard, with the best result for each metric highlighted in bold. The $\dagger$ after each metric value indicates a significant difference ($p < .05$) compared to the proposed method.}
\label{tab:Leaderboard}
    \centering
    
    \begin{tabular}{l|c|c}
        \hline
         Models & DSC (\%)  & NSD (\%) \\ 
         \hline
          Baseline & ${83.81\pm15.31}$\textsuperscript{$\dagger$}  & ${83.26\pm22.67}$\textsuperscript{$\dagger$}  \\
         LiteMedSAM-Rep~\citep{yang2024light} & $84.51\pm10.11$\textsuperscript{$\dagger$}  & $85.03\pm17.13$\textsuperscript{$\dagger$}  \\
         Rep-MedSAM \citep{wei2024rep} & $86.19\pm7.67$\textsuperscript{$\dagger$}  & $87.97\pm11.85$\textsuperscript{$\dagger$}  \\ 
         Swin-LiteMedSAM \citep{gao2024swin} & $86.78\pm8.63$\textsuperscript{$\dagger$}  & $88.44\pm12.79$\textsuperscript{$\dagger$}  \\
         MedficientSAM \citep{le2024medficientsam} & $86.20\pm8.00$\textsuperscript{$\dagger$}  & $87.65\pm11.61$\textsuperscript{$\dagger$}  \\ 
         DAFT~\citep{pfefferle2024daft} & $87.18\pm8.29$\textsuperscript{$\dagger$}  & $88.32\pm13.41$\textsuperscript{$\dagger$}  \\ 
         MCP-MedSAM (proposed) & $\mathbf{87.50\pm6.91}$& $\mathbf{89.40\pm10.37}$ \\ 
         \hline
    \end{tabular}
\end{table*}

\begin{table*}[tb]
\caption{Efficiency comparison with state-of-the-art methods on the challenge leaderboard, with the best result for each metric highlighted in bold. Notably, LiteMedSAM-Rep is trained on an NVIDIA RTX 4090, Rep-MedSAM on an NVIDIA V100, and the other models on an NVIDIA A100.}
\label{tab:Leaderboard_time}
    \centering
    \begin{tabular}{l|c|c}
        \hline
         Models &\makecell{GPU Training \\ Time (hours)}  & \makecell{CPU Inference \\ Time (seconds)} \\ 
         \hline
          Baseline  & - & - \\
         LiteMedSAM-Rep~\citep{yang2024light} & 1600.0 & 0.9 \\
         Rep-MedSAM \citep{wei2024rep} & 188.0 & 1.3 \\ 
         Swin-LiteMedSAM \citep{gao2024swin} & 106.8 & 2.6 \\
         MedficientSAM \citep{le2024medficientsam} & 118.5 & 0.7\\ 
         DAFT~\citep{pfefferle2024daft} & $>$ 42.9 & \textbf{0.4} \\ 
         MCP-MedSAM (proposed) & \textbf{23.8} & 4.6 \\ 
         \hline
    \end{tabular}
\end{table*}

\subsubsection{Ablation study}
\begin{figure*}[tbhp!]
    \centering
    \includegraphics[scale=0.25]{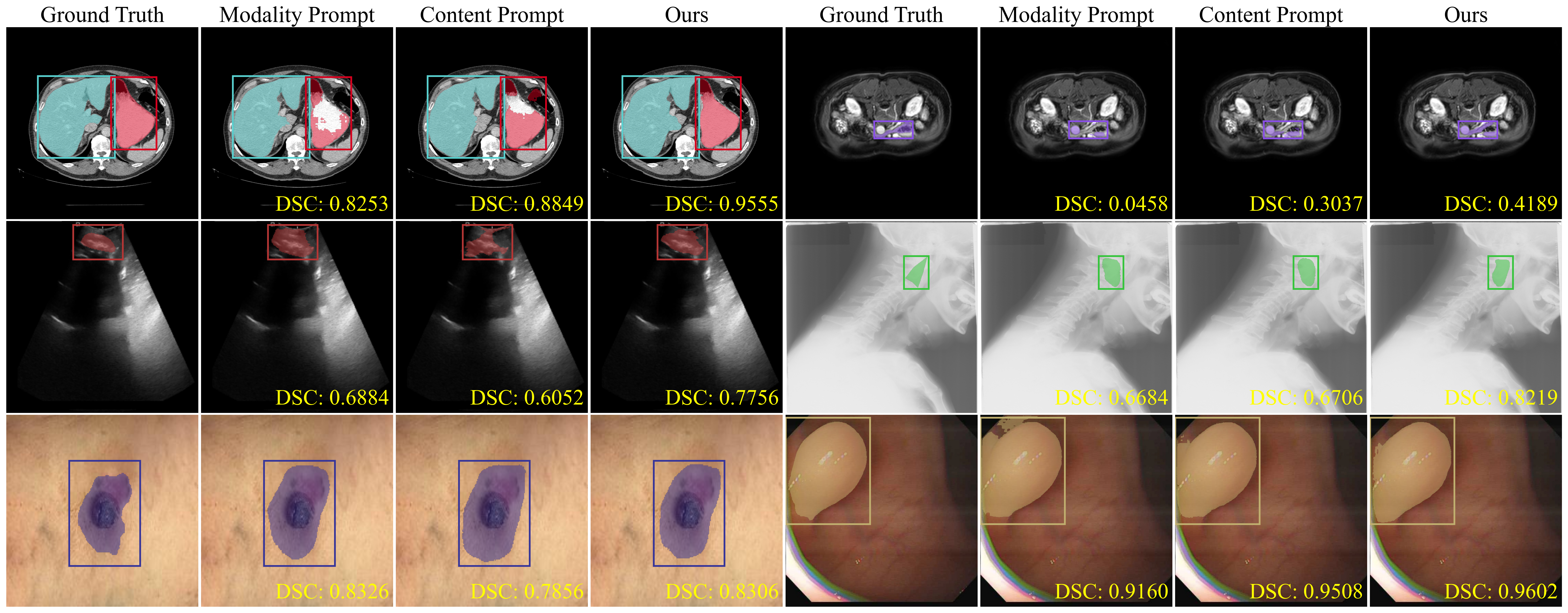}
    \caption{Visualizations generated from multiple imaging modalities using three prompting strategies: modality-only, content-only, and combined prompts. The DSC value of each sample is displayed in the bottom right corner.}
\label{fig:ab1}
\end{figure*}

In this section, we present a comprehensive ablation study of the key modules in our approach. It is mainly composed of four parts: prompt encoder components of MCP-MedSAM, pre-trained components of MCP-MedSAM, data sampling strategy and training GPU type.
\\
\\
\textbf{Prompt Encoder Components of MCP-MedSAM} Within the prompt encoder, both the modality and content prompt processing networks are composed of two key components. To assess the impact of each, we trained the model under the modality sampling strategy, omitting each component separately. Additionally, we evaluated the model using only one type of introduced prompt, as well as without any introduced prompts, to provide a comprehensive view. To note, when the CNN image encoder branch was excluded, we used a learnable matrix with randomly drawn weights from a uniform distribution as the outputting dense embedding presentation. The detailed results are shown in Table \ref{tab:prompts}. Removing a component from either introduced prompt processing network resulted in a decline in overall performance and might be even worse than removing the entire prompt processing network. Then using only the complete modality or content prompt processing network achieved comparable performance, both of which outperformed the absence of both prompts. Figure~\ref{fig:ab1} further represents the performance of each introduced prompt with some visualizations. Each prompt produced different segmentation results based on its unique perspective, with varying types of errors. Giving both prompts resulted in the best segmentation performance.
\\
\\
\textbf{Pre-trained Components of MCP-MedSAM}
As previously mentioned, we incorporated pre-trained weights into the image encoder and CLIP component of MCP-MedSAM to improve performance. To support our approach and gain deeper insights into their effects, we conducted several related ablation studies. For the tiny ViT, we tested three scenarios: without pretraining, with pre-trained weights from natural images, and with pre-trained weights from medical images. The CLIP was tested with two options: with pre-trained weights from natural images, and with pre-trained weights from medical images. As shown in Table~\ref{tab:pretrain}, both CLIP and tiny ViT had a better overall performance when using medical domain pre-trained weights compared to those initialized with natural domain pre-trained weights. Furthermore, when tiny ViT was trained from scratch without any pre-trained weights, it exhibited the lowest performance among all the options.
\\
\\
\textbf{Data Sampling Strategy}
We analyzed three different data sampling strategies: (1) Slice Sampling, used in \cite{ma2024segment}, randomly selecting a slice from the whole training dataset; (2) Case Sampling, randomly selecting a slice from each training case; (3) Modality Sampling (see algorithm \ref{alg:sampling}), introducing control over modality balance in each training batch, building on the case sampling approach. The detailed results are shown in Table \ref{tab:st}. Among them, slice sampling achieved the highest scores for CT and MR but lagged behind the other two strategies on other modalities. While case sampling and modality sampling showed stronger performance on specific modalities, modality sampling delivered the best overall results with more balanced performance across all modalities.
\\
\\
\textbf{Training GPU Type}
To further evaluate the applicability of MCP-MedSAM on smaller GPUs, we conducted an experiment on a mid-range GPU (NVIDIA RTX 6000, 24GB) by reducing the batch size from 16 to 8. The final results achieved a DSC of $86.87 \pm 7.53$ and an NSD of $88.34 \pm 12.07$, with a total training time of 54.6 hours. Overall, compared to training on an A100 GPU, segmentation performance decreased and training time increased significantly. This is mainly due to the reduced batch size and lower GPU performance. A smaller batch size leads to more iterations per epoch, resulting in longer training time. More importantly, it can also negatively affect model performance by introducing higher variance in gradient estimation, which makes training less stable and slower convergence.

\begin{table*}[tb]
\caption{Ablation study of the component of each prompt processing network in the prompt encoder part of MCP-MedSAM model. The checkmark means including the component in the model. And the best result for each evaluation metrics is shown in bold. The $\dagger$ after each metric value indicates a significant difference ($p < .05$) compared to the proposed method.}
\label{tab:prompts}
    \centering
    
    \begin{tabular}{c|c|c|c|c|c}
    \hline
    \multicolumn{2}{c|}{Modality Prompt} & \multicolumn{2}{|c|}{Content Prompt} & \multirow{2}{*}{DSC (\%) } & \multirow{2}{*}{NSD (\%) } \\
    \cline{1-4}
    Text CLIP & Modality Embedding & Image CLIP & CNN Encoder & &  \\
    \hline
         &  &  &  & $86.36\pm8.39$\textsuperscript{$\dagger$} & $87.64\pm13.23$\textsuperscript{$\dagger$} \\
          & \checkmark & \checkmark & \checkmark & $86.82\pm7.97$\textsuperscript{$\dagger$} & $88.35\pm12.72$\textsuperscript{$\dagger$} \\ 
         \checkmark & & \checkmark & \checkmark & $86.57\pm7.21$\textsuperscript{$\dagger$} & $88.39\pm10.91$\textsuperscript{$\dagger$} \\ 
         \checkmark & \checkmark & & & $87.07\pm7.19$\textsuperscript{$\dagger$} & $88.78\pm11.07$\textsuperscript{$\dagger$} \\
         \checkmark & \checkmark & & \checkmark & $86.92\pm7.83$\textsuperscript{$\dagger$} & $88.47\pm12.31$\textsuperscript{$\dagger$} \\
         \checkmark & \checkmark & \checkmark & & $86.92\pm7.59$\textsuperscript{$\dagger$} & $88.55\pm12.02$\textsuperscript{$\dagger$} \\ 
         & & \checkmark & \checkmark & $87.13\pm6.89$\textsuperscript{$\dagger$} & $88.76\pm11.12$\textsuperscript{$\dagger$} \\
         \checkmark & \checkmark & \checkmark & \checkmark & $\mathbf{87.50\pm6.91}$ & $\mathbf{89.40\pm10.37}$ \\ \hline
    \end{tabular}
\end{table*}

\begin{table*}[tb]
\caption{Ablation study of pre-trained components with different weights (natural and medical images) on the MCP-MedSAM model, with the best result for each evaluation metrics highlighted in bold. The $\dagger$ after each metric value indicates a significant difference ($p < .05$) compared to the proposed method.}
\label{tab:pretrain}
    \centering
    
    \begin{tabular}{l|c|c}
        \hline
         Method & DSC (\%)  & NSD (\%)   \\ 
         \hline
          Tiny ViT (not pre-trained)   & $85.53\pm8.67$\textsuperscript{$\dagger$} & $87.11\pm13.35$\textsuperscript{$\dagger$} \\
           Tiny ViT (pre-trained on natural images)  & $86.95\pm8.03$\textsuperscript{$\dagger$} & $88.61\pm12.36$\textsuperscript{$\dagger$} \\ 
           Tiny ViT (pre-trained on medical images, proposed)  & $\mathbf{87.50\pm6.91}$ & $\mathbf{89.40\pm10.37}$ \\
         \hline
         CLIP (pre-trained on natural images)  &$86.99\pm7.53$\textsuperscript{$\dagger$}  & $88.44\pm12.20$\textsuperscript{$\dagger$} \\ 
         CLIP  (pre-trained on medical images, proposed)  &$\mathbf{87.50\pm6.91}$ & $\mathbf{89.40\pm10.37}$ \\
         \hline
    \end{tabular}
\end{table*}

\begin{table*}[tb]
\caption{Performance comparison across different data sampling strategies, with the performance of each modality detailed. And the best performance for each modality and overall performance is shown in bold. The $\dagger$ after each metric value indicates a significant difference ($p < .05$) compared to the modality sampling strategy.}
\label{tab:st}
\centering

\resizebox{1.0\textwidth}{!}{
\begin{tabular}{l|cc|cc|cc}
\hline
\multirow{2}{*}{Modality} & \multicolumn{2}{c}{Slice Sampling} & \multicolumn{2}{c}{Case Sampling} & \multicolumn{2}{c}{Modality Sampling} \\ \cline{2-7} 
                        & DSC (\%)       & NSD (\%)       & DSC (\%)           & NSD (\%)  & DSC (\%)  & NSD (\%)    \\ \hline
CT                   &$91.00\pm9.69$ & $93.85\pm 9.70$ & $90.31\pm8.10$& $93.55\pm8.35$          & $90.02\pm7.98$& $93.43\pm8.14$\\
MR                    &$\mathbf{87.69\pm10.70}$& $\mathbf{91.68\pm12.41}$ & $85.73\pm11.07$ & $89.79\pm12.06$ & $85.53\pm11.14$& $89.81\pm12.59$\\
PET                    &$66.21\pm11.38$& $49.40\pm30.40$& $68.06\pm9.36$          & $52.05\pm29.13$          & $\mathbf{73.38\pm7.04}$ & $\mathbf{61.68\pm25.88}$ \\
US                      &$82.50\pm10.52$&$87.16\pm7.20$& $83.97\pm11.63$ & $88.71\pm8.35$ & $\mathbf{84.77\pm9.62}$ & $\mathbf{89.61\pm6.55}$ \\
X-ray                   &$83.44\pm7.19$&$88.24\pm7.36$& $\mathbf{86.33\pm5.91}$ & $\mathbf{91.00\pm6.36}$ & $85.83\pm5.93$& $90.57\pm6.27$\\
Dermoscopy             &$93.08\pm5.21$&$94.61\pm4.36$& $94.58\pm4.22$          & $96.07\pm3.23$          & $\mathbf{94.84\pm4.54}$ & $\mathbf{96.32\pm3.56}$ \\
Endoscopy               &$93.19\pm5.12$&$96.08\pm3.70$& $\mathbf{96.25\pm4.11}$ & $\mathbf{98.45\pm2.19}$ & $95.17\pm6.67$& $97.66\pm5.08$\\
Fundus                  &$94.57\pm1.70$&$96.27\pm1.52$& $95.61\pm1.41$         & $97.21\pm1.18$          & $\mathbf{95.77\pm1.39}$ & $\mathbf{97.35\pm1.21}$ \\
Microscopy              &$76.42\pm15.67$&$83.07\pm12.80$& $\mathbf{82.53\pm15.79}$          & $\mathbf{88.50\pm12.74}$         & $82.17\pm15.20$& $88.17\pm12.28$\\ \hline
Average                 &$85.34\pm8.83$\textsuperscript{$\dagger$}&$86.71\pm13.84$\textsuperscript{$\dagger$}& $87.04\pm7.97$\textsuperscript{$\dagger$}          & $88.37\pm12.87$\textsuperscript{$\dagger$}          & $\mathbf{87.50\pm6.91}$& $\mathbf{89.40\pm10.37}$ \\ \hline
\end{tabular}
    }
\end{table*}

\section{Discussion and Conclusion}
In accuracy comparisons with the benchmark models (Table~\ref{tab:Leaderboard}), MCP-MedSAM achieved the best results and showed a significant  statistical difference compared to the other methods. In general, it can be attributed to several key factors: 1) the introduction of two types of prompts offers some valuable cues for target segmentation by integrating information from different perspectives. As shown in Table~\ref{tab:prompts}, using either prompt improved overall segmentation accuracy: the modality prompt processing network captures inter-modality differences and unique target characteristics, while the content prompt processing network focuses on features within the bounding box to directly extract information about the segmentation target. The comparable performance achieved with either prompt also indicates the similar importance of modality information and content information contained in the bounding box. Moreover, the components within each prompt processing network complement each other, enriching the overall representation. In contrast, relying on a single prompt processing component may interfere with the processing of the other prompt, ultimately reducing overall performance; 2) the incorporation of pre-trained components supplies the model with prior knowledge, enhancing its final performance. In Table~\ref{tab:pretrain}, it is obvious that medical-related initial weights lead to optimal results by incorporating valuable prior information; 3) the modality-based data sampling strategy mitigates the negative effects of data imbalance, leading to a more balanced overall performance compared to the two other strategies. While this slightly lowers the performance of common modalities like CT and MRI, it significantly improves the performance of underrepresented modalities with much fewer training samples, such as PET, as shown in Table~\ref{tab:st}. Additionally, our proposed method has the lowest standard deviation values, also highlighting the effectiveness of the modality-based data sampling strategy in balancing the performance of multiple modalities. 

The visualizations also offered valuable insights for assessing the performance of MCP-MedSAM. In Figure~\ref{fig:leaderboard}, the CT and X-ray samples show multiple overlapping segmentation targets and some of them are small in size, both of which increase segmentation difficulty. In contrast, the other displayed modalities have fewer and larger targets, resulting in lower overall segmentation difficulty. However, noticeable over-segmentation occurred in Endoscopy for many benchmark models, likely due to background features resembling those of the target. They indicate that small target sizes and complex backgrounds could negatively impact segmentation performance, while the proposed MCP-MedSAM represents the best ability to mitigate these effects, demonstrating its robustness and confirming the usefulness of the previously mentioned key factors. Overall, the visualization results align with the quantitative findings in Table~\ref{tab:Leaderboard}. Likewise, Figure~\ref{fig:ab1} more clearly demonstrates the benefits of incorporating both introduced prompts into the prompt encoder, as their combination further enriches the feature representation, leading to enhanced performance.

In efficiency comparisons with the benchmark models (Table~\ref{tab:Leaderboard_time}), MCP-MedSAM required the least training time, while the other models took significantly longer time to finish training. For the inference time, the main goal of MCP-MedSAM is to achieve superior segmentation performance without requiring long training time and significant GPU resource consumption, so improving inference speed is not our primary focus. Furthermore, incorporating additional components such as the CLIP component and CNN image encoder will inevitably increase inference time. Several of the benchmark models have adopted different strategies to reduce inference time while maintaining performance, for example, DAFT~
\citep{pfefferle2024daft} replaced PyTorch with the OpenVINO Runtime\footnote{https://docs.openvino.ai/2024/openvino-workflow/running-inference.html}. Similarly, Medficientsam~\citep{le2024medficientsam} further optimized inference by integrating OpenVINO and leveraging C++ for pre-processing and post-processing optimizations. It could be interesting to adopt such strategies for MCP-MedSAM as well, which will be considered in our future work. Nonetheless, for most clinical scenarios, inference times in the range of seconds are very acceptable, while they can be further reduced by running the models on a GPU.

MCP-MedSAM features a lightweight structure compared to MedSAM \citep{ma2024segment}. While a direct performance comparison is not feasible due to differences in training and testing datasets, the reported performance metrics, such as DSC (ranging from 0.85 to 0.90), indicate comparable performance levels. This highlights MCP-MedSAM's ability to achieve effective medical image segmentation despite its lightweight design. These findings underscore the potential to deliver high-quality segmentation performance without extensive GPU resources, encouraging the development of lightweight and accessible models for advanced medical image analysis. However, MCP-MedSAM adopts the overall design of MedSAM and segments 3D data slice by slice in a 2D manner. This approach is not only time-consuming but also negatively impacts the segmentation performance of 3D samples, as it prevents the model from learning the correlations between slices. With the proposal of the SAM2 framework \citep{ravi2024sam}, which achieves superior performance in segmenting both images and videos, integrating SAM2's working principles with the MCP-MedSAM structure has the potential to enhance segmentation performance for 3D modalities, while also reducing overall inference time. Furthermore, although our results clearly demonstrate quantitative improvement in segmentation performance, further investigation is still needed to explore how these enhancements translate into practical clinical value. For example, in radiotherapy (RT) planning, the segmentation of targets such as tumors is critical, as even small errors can significantly impact dose distribution and treatment effectiveness. Therefore, DSC improvements in tumor segmentation are clinically meaningful.
However, for certain organs-at-risk (OARs) that are located far from the tumor region—such as the esophagus in some head and neck cancer cases, \cite{mody2024large} showed a low correlation between DSC and dose errors. In such cases, a marginal gain in DSC may not translate into a noticeable clinical difference. Some future studies can further investigate the real-world impacts of these segmentation improvements across various clinical scenarios.

In this work, we proposed a lightweight medical segment anything model called MCP-MedSAM, designed to achieve strong overall performance without long training time and large GPU resource consumption. By integrating pre-trained components, the model training process is accelerated, leading to improved performance. Then the introduction of the modality prompt and the content prompts offers valuable diverse information, improving upon the lightweight MedSAM design. Furthermore, a modality-based data sampling strategy ensures that each modality is trained equally, leading to a more balanced overall performance. In conclusion, MCP-MedSAM achieves better overall segmentation performance against top-ranking methods in the challenge\textsuperscript{\ref{comp}}, demonstrating its effectiveness and potential.

\acks{This study was supported by the China Scholarship Council (No. 202207720085) and the project ROBUST: Trustworthy AI-based Systems for Sustainable Growth with
project number KICH3.LTP.20.006, which is (partly)
financed by the Dutch Research Council (NWO), Philips Research, and the Dutch Ministry of Economic Affairs
and Climate Policy (EZK) under the program LTP KIC
2020-2023. This study utilized the Dutch national e-infrastructure with the support of the SURF Cooperative using grant No. EINF-6458.}

\ethics{The work follows appropriate ethical standards in conducting research and writing the manuscript, following all applicable laws and regulations regarding treatment of animals or human subjects.}

\coi{We declare we don't have conflicts of interest.}

\data{All data used in the experiments is publicly available.}

\bibliography{ref}






\end{document}